\documentclass[preprint,12pt]{elsarticle}

\usepackage{amssymb}
\usepackage{amsmath}
\usepackage{mathrsfs}
\usepackage{graphicx}
\usepackage{subcaption}
\usepackage{url}
\usepackage{hyperref}
\usepackage{verbatim}
\hypersetup{
    colorlinks=true,
    linkcolor=blue,
    filecolor=blue,      
    urlcolor=blue,
    citecolor=cyan,
}

\makeatletter
\newif\if@restonecol
\makeatother

\usepackage[linesnumbered,ruled,vlined]{algorithm2e}%[ruled,vlined]{
\usepackage{algpseudocode}
\usepackage{amsmath}
\journal{Knowledge-Based Systems}

\begin{document}

\begin{frontmatter}

%% Title, authors and addresses

%% use the tnoteref command within \title for footnotes;
%% use the tnotetext command for theassociated footnote;
%% use the fnref command within \author or \address for footnotes;
%% use the fntext command for theassociated footnote;
%% use the corref command within \author for corresponding author footnotes;
%% use the cortext command for theassociated footnote;
%% use the ead command for the email address,
%% and the form \ead[url] for the home page:
%% \title{Title\tnoteref{label1}}
%% \tnotetext[label1]{}
%% \author{Name\corref{cor1}\fnref{label2}}
%% \ead{email address}
%% \ead[url]{home page}
%% \fntext[label2]{}
%% \cortext[cor1]{}
%% \address{Address\fnref{label3}}
%% \fntext[label3]{}

%\title{DNNRE: A Dynamic Neural Network for Distant Supervised Relation Extraction}
\title{Improving Distant Supervised Relation Extraction by Dynamic Neural Network}

%% use optional labels to link authors explicitly to addresses:
%% \author[label1,label2]{}
%% \address[label1]{}
%% \address[label2]{}

\author[scu_ee]{Yanjie~Gou}
\ead{yanjie.gou@stu.scu.edu.cn}

\author[scu_ee]{Yinjie~Lei\corref{cor}}
\ead{yinjie@scu.edu.cn}

\author[ua]{Lingqiao~Liu}
\ead{lingqiao.liu@adelaide.edu.au}

\author[dut]{Pingping~Zhang}
\ead{jssxzhpp@mail.dlut.edu.cn}

\author[scu_cs]{Xi~Peng}
\ead{pengx.gm@gmail.com}

\address[scu_ee]{College of Electronics and Information Engineering, Sichuan University, China}
\address[ua]{School of Computer Science, The University of Adelaide, Australia}
\address[dut]{School of Information and Communication Engineering, Dalian University of Technology, China}
\address[scu_cs]{College of Computer Science, Sichuan University, China}

%\fntext[fn1]{The second author has the equal contribution as the first author for this work.}

\cortext[cor]{Corresponding author}

%\author{Yanjie Gou, Yinjie Lei, Lingqiao Liu, \\ Pingping Zhang, Xi Peng}
%\address{Sichuan University, China}
%\address{The University of Adelaide, Australia}
%\address{Dalian University of Technology, China}

\begin{abstract}
Distant Supervised Relation Extraction (DSRE) is usually formulated as a problem of classifying a bag of sentences that contain two query entities, into the predefined relation classes. Most existing methods consider those relation classes as distinct semantic categories while ignoring their potential connection to query entities. 
In this paper, we propose to leverage this connection to improve the relation extraction accuracy. Our key ideas are twofold:
(1) For sentences belonging to the same relation class, the expression style, i.e. words choice, can vary according to the query entities. To account for this style shift, the model should adjust its parameters in accordance with entity types. 
(2) Some relation classes are semantically similar, and the entity types appear in one relation may also appear in others. Therefore, it can be trained cross different relation classes and further enhance those classes with few samples, i.e., long-tail classes.  
To unify these two arguments, we developed a novel \textbf{D}ynamic \textbf{N}eural \textbf{N}etwork for \textbf{R}elation \textbf{E}xtraction (DNNRE). The network adopts a novel dynamic parameter generator that dynamically generates the network parameters according to the query entity types and relation classes. By using this mechanism, the network can simultaneously handle the style shift problem and enhance the prediction accuracy for long-tail classes.
Through our experimental study, we demonstrate the effectiveness of the proposed method and show that it can achieve superior performance over the state-of-the-art methods.
\end{abstract}

%%Graphical abstract
%\begin{graphicalabstract}
%\includegraphics{grabs}
%\end{graphicalabstract}

%%Research highlights
%\begin{highlights}
%\item Research highlight 1
%\item Research highlight 2
%\end{highlights}

\begin{keyword}
%% keywords here, in the form: keyword \sep keyword
Deep neural network \sep Distant supervision \sep Relation extraction \sep Dynamic paremater \sep Style shift \sep Long-tail relation
%% PACS codes here, in the form: \PACS code \sep code

%% MSC codes here, in the form: \MSC code \sep code
%% or \MSC[2008] code \sep code (2000 is the default)

\end{keyword}

\end{frontmatter}

%% \linenumbers

%% main text
%\section{}
%\label{}
\section{Introduction} \label{intro}

\begin{figure}[htb]
\begin{center}
\includegraphics[width=0.8\textwidth]{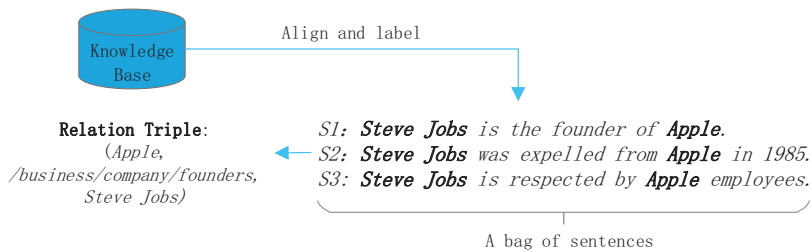}
\caption{\label{align} An example of how the training instances are generated through distant supervision. \textit{S2} and \textit{S3} are wrong labeled with the relation ``\textit{/business/company/founders}'' since they do not express the relation.}
\end{center}
\end{figure}

Relation Extraction (RE) \cite{zelenko2003kernel, mooney2006subsequence} aims to extract relations of entities from sentences, which can automate the construction of Knowledge Bases (KBs) and has potential benefits to downstream applications such as question answering \cite{sadeghi2015viske,ravichandran2002learning} and web search \cite{yan2009unsupervised}. Due to the difficulty of collecting a large amount of sentence-level annotations, most recent RE methods are based on the Distant Supervision (DS) framework \cite{mintz2009distant} which can automatically annotates adequate amounts of data by align texts with KBs: for a bag of sentences that contains two entities, if KBs has this entity pair, this bag will be labeled with the corresponding relation of the entity pair in KBs.
With the DS framework, RE can be cast as a problem of classifying a bag of sentences which contain the same query entity pair, into predefined relation classes. 
However, DS framework has its own disadvantages, that is, the noise label and long-tail problems. 
Figure \ref{align} shows how the label of a sentence bag is annotated and how the noise label problem is introduced by DS framework. Table \ref{longtail_num} shows the long-tail problem existing on the NYT dataset, which is a widely used DSRE dataset \cite{riedel2010modeling}.

\begin{table}[htb]
\caption{\label{longtail_num}The sample numbers of different relation classes in the NYT training set, which is pretty class-imbalanced: the samples of half relation classes is less than 100.}
\centering	
\begin{tabular}{cc}
\hline 
realtion class & number of class samples\tabularnewline
\hline 
\textit{/location/location/contains} & 75969\tabularnewline
\textit{/people/person/nationality} & 11446\tabularnewline
\textit{/business/person/company} & 7987\tabularnewline
\textit{/location/us\_state/capital} & 798\tabularnewline
\textit{/business/company/place\_founded} & 677\tabularnewline
... & ...\tabularnewline
\textit{/location/it\_region/capital} & 22\tabularnewline
\textit{/business/company/locations} & 19\tabularnewline
\textit{/broadcast/content/location} & 8\tabularnewline
\textit{/location/jp\_prefecture/capital} & 2\tabularnewline
\textit{/business/shopping\_center/owner} & 1\tabularnewline
\hline 
\end{tabular}
\end{table}

In the existing works, many efforts \cite{riedel2010modeling,surdeanu2012multi,zeng2015distant} have been devoted to reducing the effect of the noise label problem by recognizing the valid sentences. To be specific, the attention mechanism \cite{lin2016neural}, its variants \cite{du2018multi,yuan2019cross,huang2019self} and other methods \cite{liu2017soft, wu2017adversarial} are proposed to address the noise label problem and achieve new state-of-the-arts.
However, these works consider relations in isolation and ignore the connections between relation classes.
To take advantage of those connections, 
\cite{ye2017jointly} utilizes class ties in relations and incorporate it with a pairwise ranking framework.
\cite{han2018hierarchical} utilizes the relation hierarchies in KBs and proposes a hierarchical attention scheme to further enhance the relation representations.
Based on \cite{han2018hierarchical}, \cite{katt_naacl19} integrates the Knowledge Graph Embedding into relation extraction model and improves the long-tail class accuracy. 
They both transfer the well-trained features to the long-tail classes by connecting relations according to the KB hierarchies. 
Besides, there are also works utilize external information (e.g., entity type information \cite{liu2014exploring,vashishth2018reside}) to enhance DSRE. They consider the entity type information as input features, which can be classified along with the sentence features. Nevertheless, the way these approaches \cite{liu2014exploring,vashishth2018reside} utilizing the entity type information is inefficient, since they ignore the connections between relation classes and query entities.

\begin{figure}[htb]
\begin{center}
\includegraphics[width=0.9\textwidth]{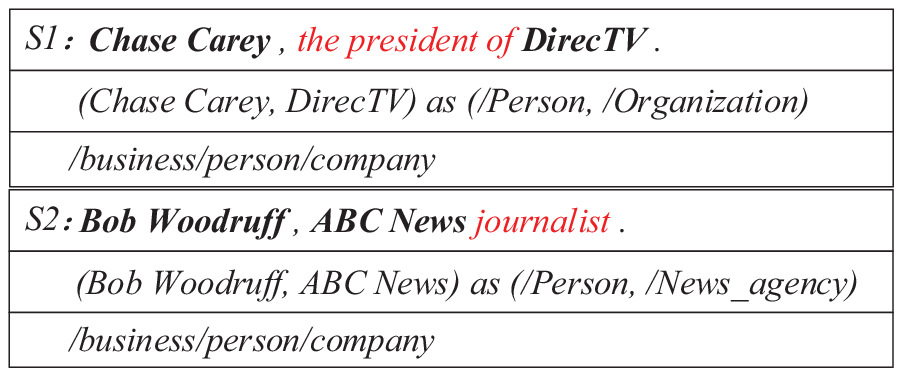}
\caption{\label{subclass} An example of the style shift problem in DSRE. The keywords that convey this relation are in red font.
}
\end{center}
\end{figure}

This paper studies the DSRE problem by re-examining the relation definitions in the existing methods. On one hand, the class definitions may not be fine-grained enough, since the style that a sentence expresses the entity relation may vary for different query entity pairs. For example, given two sentences in Figure \ref{subclass}, they both express the same relation. However, the keywords that convey this relation are quite different due to the difference in query entity types, i.e., (\textit{Person, Organization}) VS. (\textit{Person, News agency}). 
It seems that we need to further consider the style shift problem of each relation class concerning the entity types. 

On the other hand, some relation definitions may be fine-grained since many relations are semantically related and their representations may be similar in the feature space. For example:
\begin{itemize}
\item ``\textit{/people/person/place\_of\_birth}'' and ``\textit{/people/person/place\_lived}''\\ both indicate someone appears somewhere at a certain time. 
\end{itemize}
Moreover, the head and tail entities mentioned in these two relation classes are with the ``\textit{/person}'' and ``\textit{/location}'' types, respectively. That is, they can be trained well cross different relation bags. 
It can be expected that based on this intuition, the well-trained entity type information can be utilized to further enhance the vanilla relation representations. That especially benefits the long-tail problem.

To unify these two arguments, we propose to use a dynamic neural network in which the class-dependent parameters (i.e., attention and classifier) can be dynamically determined by the query entity type information.
By doing so, we can make our prediction model adaptive to the query entity types which can naturally deal with the style shift problem. Also, the long-tail relation representations can be derived from those entity type information, and it can further address the long-tail problem. By utilizing the type information this way, we can  tackle DSRE more efficiently.

Specifically, to dynamically generate class-dependent parameters (i.e., attention and classifier) for our model, we develop a novel dynamic parameter generation module. Such a module generates network parameters in two steps: (1) The corresponding entity types of one bag are firstly converted to type embeddings. Then, a relation-aware attention mechanism\footnote{One entity may have multiple entity types. Therefore, the information of the entity types needs to be aggregated effectively.} is proposed to aggregate the important information of them for a specific relation class. 
That is, different relations may correspond to different entity types. 
(2) The aggregated information of entity types is incorporated with the vanilla relation representations, and it is further transformed into the dynamic class representations by nonlinear mapping.

We conduct experiments on a widely used large-scale DSRE benchmark dataset, and the experimental results demonstrate the superior performance of the proposed method. It is validated that the dynamic network design is beneficial for handling both style shift and long-tail problems in DSRE. 

In summary, our main contributions of this work are as follows:
\begin{itemize}
    \item We first utilize the class relationship with entity types for improving the performance of DSRE.
    \item We propose a novel dynamic parameter generator to build a dynamic neural network whose parameters are determined by the query entity types and relation classes.
	 \item We propose a novel relation-aware attention over entity types to aggregate the discriminative information in the entity types.
    \item Our experiments on a widely used benchmark show that our method gives a new state-of-the-art result.
\end{itemize}

\section{Related Works}

\subsection{Hand-crafted Feature Based Methods}

In its early years, most of the DSRE methods are based on the hand-crafted features \cite{mintz2009distant,riedel2010modeling,hoffmann2011knowledge}, e.g., POS tags, named entity tags, and dependency paths.
\cite{mintz2009distant} assumes that sentences containing the same entity pair, all express the same relation. However, this assumption does not always hold. 
To relax this assumption, \cite{riedel2010modeling} assumes that if two entities are held in a relation, at least one sentence mentioning these entities may express such relation. Then, they employ the multi-instance learning (MIL) paradigm to support this assumption. 
Later, since different relational triplets may have overlaps in a sentence and MIL can not address this problem, \cite{hoffmann2011knowledge,surdeanu2012multi} extend MIL to multi-instance multi-label paradigm to handle this problem. However, the hand-crafted features are not sufficiently robust, as well as lead to the error propagation problem.

\subsection{Deep-feature Based Methods}

In recent years, researchers turn to apply deep learning to DSRE due to its promising performance and generalization ability in various NLP applications. Many methods \cite{zeng2015distant,lin2016neural,ji2017distant} are under the MIL paradigm framework aiming to denoise the wrong label generated by DS. 
\cite{zeng2015distant} proposes a piecewise convolutional neural network, and utilizes MIL to select one sentence in a bag which can well express the relation between the entity pair within such a sentence. 
However, they omit useful information in other sentences, which are also useful for expressing such a relation. To solve this problem, \cite{lin2016neural} introduces a selective attention mechanism to capture more useful information in other sentences. 
Afterwards, its variants \cite{du2018multi,yuan2019cross,huang2019self,ye-ling-2019-distant} are proposed, 
where \cite{du2018multi} introduces a multi-lingual neural relation extraction framework to better utilize the information from various languages; 
\cite{ye-ling-2019-distant} considers both intra-bag and inter-bag attentions in order to deal with the noise at sentence-level and bag-level respectively;
\cite{yuan2019cross} extends the selective attention to cross-relation cross-bag selective attention and trains the model more noise-robust; 
\cite{huang2019self} enhances convolutional neural network with the self attention mechanism to learn better sentence representations;

Besides, \cite{zheng2017joint} proposes a tagging based method that jointly extracts entities and relations. 
\cite{ijcai2018-620} converts the joint task into a directed graph by designing a graph scheme and propose a transition-based approach to generate the directed graph.
\cite{liu2018neural} reduces the noise via sub-tree parse and utilizes transfer learning to improve the performance of DSRE. 
\cite{su2018globalR} proposes to embed textual relations with global statistics of relations
\cite{han2018neural} proposes a joint representation learning framework on knowledge graph completion and relation extraction from text. 
\cite{he2018see} proposes to learn syntax-aware entity embedding for DSRE. 
\cite{jia2019arnor} proposes an attention regularization method to further reduce the noise in the dataset. \cite{zhengetal2019diag} proposes a neural pattern diagnosis framework, which can automatically summarize and refine high-quality relational patterns from noisy data.
Other learning strategies, like adversarial training \cite{wu2017adversarial,wang2018adversarial,qin-etal-2018-dsgan}, capsule network \cite{zhangetal2018attention,zhang2019multi}, and reinforcement learning \cite{feng2018reinforcement,takanobu2019hierarchical} are also applied to DSRE to further improve its performance. 

Recently, due to the powerful capabilities of the pre-training language model, \cite{altetal2019fine} utilizes GPT \cite{radford2018improving} for DSRE and predicts a larger set of distinct relation types with high confidence. In another way, \cite{chenetal2019global} studies how to learn a general-purpose embedding of textual relations and further improves the DSRE performance.

\subsection{Methods Incorporating External Information}

Recently, other useful external information is identified to be beneficial for DSRE, e.g., KB information. 
\cite{ji2017distant} utilizes entity descriptions for DSRE, which can provide rich background information of entities, and help recognize relations in DSRE. 
\cite{liu2014exploring} explores fine-grained entity type constraints for DSRE. 
\cite{liu2017heterogeneous} utilizes annotations from the heterogeneous information sources, e.g., knowledge base and domain heuristics, and conduct relation extractor learning.
\cite{vashishth2018reside} uses a set of side information, e.g., entity type, and relation alias, to boost DSRE performance. 
\cite{wang-etal-2018-label-free} proposes a label-free distant supervision method, and only uses the prior knowledge derived from the KG to supervise the learning of the classifier softly.
\cite{lei2018cooperative} leverages the corpus-based and KG-based information, and use logic rules on the entity type level. 
\cite{han2018hierarchical} proposes a coarse-to-fine grained attention scheme by hierarchical relation structures in KB. 
Based on \cite{han2018hierarchical}, \cite{katt_naacl19} proposes a knowledge-aware attention scheme using Knowledge Graph embedding (KGE). 
Besides, \cite{Beltagy2019Comb} combines the distant supervision data with additional directly-supervised data to train a model for identifying valid sentences. 
\cite{hu2019improving} proposes a multi-layer attention-based model to improve DSRE with joint label embedding, which is obtained from entity descriptions and KG.

However, all the above works ignore the style shift problem, whereas DNNRE uses the entity type information to address it and further improve DSRE performance. 
Note that there are also previous works using entity types in their models \cite{vashishth2018reside}. However, they are quite different from us: we utilize entity types to dynamically generate the parameters in our model for addressing the style shift and long-tail problem, whereas previous works just use entity types as input features.

\begin{figure}[htb]
\begin{center}
\includegraphics[width=1.\textwidth]{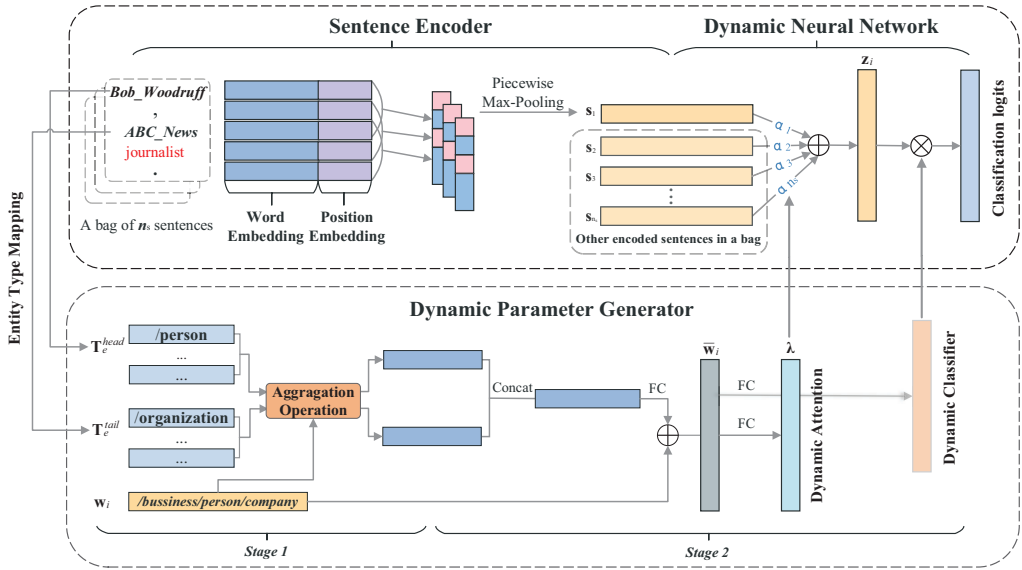}
\caption{\label{main_framework}Overview of \textbf{DNNRE}. The sentence encoder is in the top left. The dynamic Neural network parts are in the top right, in which the attention (Dynamic Attention) and classifier (Dynamic Classifier) are generated by the dynamic parameter generator in the bottom. Note that the fully connected layers (FC) which map $\overline{w}_i$ to dynamic attention and classifier modules \textbf{do not} share parameters.}
\end{center}
\end{figure}

\section{Methodology}

The primary idea of the proposed method is to build a network with \textbf{DYNAMIC} weights, that is, parts of the network parameters will be dynamically generated by the entity types and relation classes. This is in contrast to the traditional methods which use \textbf{STATIC} models for which the model parameters will be fixed during testing. Formally, the class-dependent parameters $\boldsymbol{\lambda}$ of the proposed network can be dynamically generated by the following function:
\begin{equation}\label{eq:overall_idea}
    \boldsymbol{\lambda} = \phi(\mathbf{T}_{e}, \mathbf{w}_i),
\end{equation}
where $\mathbf{T}_{e}$ (i.e., $\mathbf{T}_e^{head}$ and $\mathbf{T}_e^{tail}$) is the entity types, 
and $\mathbf{w}_{i}$ is the vanilla relation representation of $i$-th class in the attention or classifier modules.
The function $\phi$ is called \textbf{dynamic parameter generator} which transfer $\mathbf{T}_{e}$ and $\mathbf{w}_i$ into the network parameters $\boldsymbol{\lambda}$.

Since $\mathbf{T}_{e}$ is a variable of the query entity types, the generated network parameters will be online adapted at the test stage, which offers a solution to compensate for the style shift and long-tail problems. 

\subsection{Overall Architecture}

The overall architecture of the proposed DNNRE is illustrated in Figure \ref{main_framework}. In the top right, the \textbf{Sentence Encoder} encodes a bag of sentences into sentence representations.
Meanwhile, the \textbf{Entity Type Mapping} maps the entity pair to $\mathbf{T}_{e}$ for enhancing the relation representations. 
Then, this information will be utilized to generate the dynamic parameters by the \textbf{Dynamic Parameter Generator} (bottom). Finally, the dynamic parameters will build the \textbf{Dynamic Neural Network} (i.e., dynamic attention and classifier) in the top right. The \textbf{Dynamic Attention} aggregates the sentence representations into a bag representation, which is feed into the \textbf{Dynamic Classifier} to predict its corresponding relation class.

In the remaining of this section, the contents are organized as follows:

\begin{itemize}
    \item Firstly, the Sentence Encoder will be introduced briefly in subsection \ref{encoder}.
    \item Then, the Entity Type Mapping (i.e., $\mathbf{T}_{e}$) will be introduced in subsection \ref{input_des}.
    \item The Dynamic Parameter Generator will be elaborated in subsection \ref{generator}.
    \item Finally, the Dynamic Neural Network  is introduced in subsection \ref{dyna_parts}.
\end{itemize}

\subsection{Sentence Encoder}\label{encoder}

In the framework of DSRE, the input of the network is a bag of sentences. Similar as most DSRE methods \cite{zeng2015distant,lin2016neural}, we first convert each sentence $\mathscr{\mathscr{\mathcal{S}_i}}=\left\{ w_{1},w_{2},...,w_{|s|}\right\}$ into a fixed-length vector $\mathbf{s}_i$ by using a sentence encoder. In this work, we use PCNN \cite{zeng2015distant} to fulfill this task. The sentence encoder has three layers: input representation layer, encoding layer and pooling layer, which are introduced as follows:

\subsubsection{Input Representation Layer}

The input representation layer converts each word in a sentence by the word embedding and the position embedding.

The word embedding is used to represent each word token $i$ of $w_{i}$ by a pre-trained word embedding vector $\mathbf{v}_{i}$, which is trained on NYT corpus by the word2vec tool\footnote{https://code.google.com/p/word2vec/} \cite{mikolov2013efficient}.

The position embedding consists of two fixed dimension vectors for representing the relative positions between $w_{i}$ and entity pair. We concatenate position embedding $\mathbf{p}_{i1},\mathbf{p}_{i2}$ to the word representation. 

\subsubsection{Encoding Layer}
The word representations $\mathbf{x}_{i}=\left[\mathbf{v}_{i};\mathbf{p}_{i1};\mathbf{p}_{i2}\right]\in \mathbb{R}^{d_{i}}$ ($d_{i}=d_{w}+2\times d_{p}$) are fed into the encoding layer. $m$ convolution kernels $\mathbf{K}=\left\{ \mathbf{k}_{1},...,\mathbf{k}_{m}\right\} \in \mathbb{R}^{n_{w}\times d_{i}}$ slide over the input to capture features in the ${n_{w}}$-gram:
\begin{equation}\label{eq1}
    \mathbf{h}_{i}=\mathbf{k}_{i}*\mathbf{x}_{j-n+1:j}\quad1\leq i\leq m,
\end{equation}
where $\mathbf{x}_{j-n+1:j}$ means the word representation from index $j-n+1$ to $j$. Afterwards, we can obtain $\mathbf{H} = \left\{ \mathbf{h}_{1},...\mathbf{h}_{m}\right\}$.

\subsubsection{Pooling Layer}
After this convolution operation, a piecewise max-pooling is adopted to aggregate word-level information. Supposed $\mathbf{h}_{i}$ is split into $\left\{\mathbf{h}_{i1},\mathbf{h}_{i2},\mathbf{h}_{i3}\right\}$ by the entity positions, this pooling method is described as below:
\begin{equation}\label{eq2}
    \mathbf{q}_{i}=[maxpool(\mathbf{h}_{ij})]\quad j=1,2,3.
\end{equation}

Then we obtain $\mathbf{Q} \in \mathbb{R}^{m\times 3}$, and $\mathbf{Q}$ is flattened to a vector and translate it into the sentence embedding $\mathbf{s} \in \mathbb{R}^{d_{s}}$ by a non-linear layer.

\newcommand{\tabincell}[2]{\begin{tabular}{@{}#1@{}}#2\end{tabular}}  
\begin{table}[htb]
\caption{\label{typeset}The entity types used in our model. The left column lists coarse-grained entity types and the right column lists fine-grained entity types corresponding to the left. We put the hard-categorized entity types in the bottom. Note that we only show part of the 112 entity types due to the space constraint.}
\begin{center}
\begin{tabular}{c|c}
\hline 
Coarse-grained&Fine-grained\tabularnewline
\hline 
person & \tabincell{c}{doctor engineer architect coach actorreligious\_leader\\monarch terrorist  artist musician athlete director}\tabularnewline %\\politician soldier author
\hline
organization & \tabincell{c}{terrorist\_organization airline government company\\government\_agency educational\_institution military\\fraternity\_sorority political\_party news\_agency}\tabularnewline %\\sports\_league sports\_team educational\_department
\hline 
location & \tabincell{c}{body\_of\_water city island country park astral\_body\\mountain county glacier province bridge road}\tabularnewline %\\railway cemetery
\hline 
product & \tabincell{c}{camera engine mobile\_phone airplane software  weapon\\computer ship game spacecraft instrument train car\\}\tabularnewline
\hline 
art & \tabincell{c}{written\_work film newspaper play music}\tabularnewline
\hline 
event & \tabincell{c}{military\_conflict attack natural\_disaster election\\sports\_event protest terrorist\_attack}\tabularnewline
\hline 
building & \tabincell{c}{airport dam hospital hotel library power\_station\\ restaurant sports\_facility theater}\tabularnewline
\hline 
\multicolumn{2}{c}{\tabincell{c}{chemical\_thing website color biological\_thing award time disease god\\broadcast\_network medical\_treatment broadcast\_program title drug\\educational\_degree tv\_channel symptom currency law algorithm\\ethnicity body\_part language living\_thing food animal religion\\programming\_language stock\_exchange transit\_system transit\_line}}\tabularnewline
\hline 
\end{tabular}
\end{center}
\end{table}

\subsection{External Information Acquisition} \label{input_des}
%\subsection{Entity Type Mapping} \label{input_des}

The external information of entity types has been proved to be useful for the DSRE task as additional input features \cite{liu2014exploring,vashishth2018reside}. Unlike those existing works, we use the entity type information $\mathbf{T}_{e}$ to dynamically determine the network parameters. The representation of it is shown as follows. 

We first obtain the entity types from KB and further mapping to the types predefined by FIGER \cite{ling2012fine}, which are shown in Table \ref{typeset}. 
Then, we create an embedding vector for each entity type. Note that in practice, one entity may correspond to multiple entity types, in such a case, we propose a relation-aware attention to selectively aggregate the important information in these entity type embedding and obtain an aggregated vector to represent them. Besides, some entities may correspond to no entity type, we then use an ``\textit{UNK}'' token to represent their entity type.

Note that if the entity type information can not be directly obtained from KBs, we can also use the entity typing methods \cite{ling2012fine,xin2018improving} to obtain this information. In this paper, we assume the entity type information is easy to access and directly obtain it from KBs.

\subsection{Dynamic Parameter Generator} \label{generator}
In the following, the implementation of the dynamic parameter generator $\phi(\mathbf{T}_{e}, \mathbf{w}_i)$ is elaborated in detail. In our design, it achieves parameter generation through two stages: 
(1) In \textit{stage 1}, since one entity may correspond to several entity types, we aggregate the optimal information in these entity types which is most important to enhance the relation representations. (2) In \textit{stage 2}, the aggregated entity type information is further utilized to generate the dynamic parameters.

\subsubsection{Stage 1: Type Information Aggregation}
The first stage is to aggregate the type information $\mathbf{T}_{e}$ (i.e., $\mathbf{T}_e^{head}$ and $\mathbf{T}_e^{tail}$) into two embedding vectors $\mathbf{t}_e^{h}$ and $\mathbf{t}_e^{t}$ with the dimension of $d_n$, which represents the aggregated head entity and tail entity type information, respectively. 

The reason to selectively aggregate such type information is that different relation classes may correspond to different entity types. For example: the location ``Sacramento'' can be mentioned by relations ``\textit{/location/us\_state/capital}'' and ``\textit{/people/person/place\_of\_birth}''. The former may attend more to its type ``\textit{/city}'', and the latter may attend more to ``\textit{/location}''.

This step is achieved by a relation-aware attention over the entity type embedding. This attention mechanism enables the model to aggregate the most discriminative information related to a specific relation class. The attention weights over the type embedding are calculated as follows:
\begin{align}\label{eq3}
    &\alpha_{i} = \mathbf{t}_{i}^h\mathbf{W}_{t}{\mathbf{w}}_k^T, \nonumber \\
	&\mathbf{t}_e^{h} = \sum_{i=1}^{n_{s}}\frac{\exp(\alpha_{i})}{\sum_{j=1}^{n_s} \exp(\alpha_{j}) }\mathbf{t}_{i}^h,
\end{align}
where $\mathbf{W}_{t} \in \mathbb{R}^{{d_{t}}\times{d_{r}}}$ are learnable parameters. $\mathbf{w}_{k} \in \mathbb{R}^{d_{r}}$ is the static representation for the $i$-th class in the vanilla attention or classifier. $\mathbf{t}_{i}^h$ is one of the type embedding vectors in $\mathbf{T_e^{head}}$, which is a set of entity type embeddings $\left\{ \mathbf{t}_{1}^h,...,\mathbf{t}_{|T_{head}|}^{h}\right\}$. $\mathbf{t}_e^{t}$ is also aggregated from $\mathbf{T_e^{tail}}$ by the same aggregation strategy shown above.

Note that we also investigate another two aggregation strategies (i.e., average pooling and max pooling) to compare with the proposed relation-aware attention over the entity types. One can refer to subsection \ref{aggregation_strategy} for a detailed discussion.

\subsubsection{Stage 2: Dynamic Parameters Generation}

After we have aggregated the entity type information and obtained $\mathbf{t}_e^{h}$ and $\mathbf{t}_e^{t}$, they are further utilized to generate the dynamic class parameters by the following transformation:
\begin{equation}
    \overline{\mathbf{w}}_{k}=f_{d}(\mathbf{w}_{k}+f_{t}([\mathbf{t}_e^{h},\mathbf{t}_e^{t}])),
\end{equation}
where the term $f_{t}([\mathbf{t}_e^{h},\mathbf{t}_e^{t}])$ can be considered as a dynamic component generated from the information of the head and tail entity types. $f_{d}(\cdot)$ and  $f_{t}(\cdot)$ both denote the fully connected layers, and two layer fully connected module is applied in our design. Besides, $[\cdot, \cdot]$ denotes concatenation of two vectors.

After the transformation, $\overline{\mathbf{w}}_{k}$ can be utilized as the network parameters of the $k$-th class in the attention and the classifier. In our work, the dynamic parameters of the attention and classifier used two different $f_d(\cdot)$. That is, the Dynamic Parameter Generator for the attention and classifier share its parameters except for the fully connected module $f_d(\cdot)$.
For clarity, we denote them as $\phi_a(\mathbf{T}_{e}, \mathbf{r}_k) \in \mathbb{R}^{d_r}$ and $\phi_c(\mathbf{T}_{e}, \mathbf{w}_k) \in \mathbb{R}^{d_r}$, where $\mathbf{r}_k, \mathbf{w}_k$ are the vanilla static parameters for the $k$-th class in the attention and classifier, respectively. $d_r$ is the dimension of the parameters for each class. 

\subsection{Dynamic Neural Network} \label{dyna_parts}

After the sentences being encoded into vector representations, the next operation is to aggregate them into a bag representation by attention mechanism. Finally, the bag representation is fed into a classifier. 

Attention and classifier both measure the similarity between features and relations, at sentence and bag level, respectively.  In that sense, the dynamic parameter generator can enhance both of them, which will be introduced as dynamic attention and dynamic classifier module in the following parts.

\subsubsection{Dynamic Attention}

Given $n_{s}$ sentences in a bag, their corresponding features are extracted by PCNN as $\mathbf{S}=\{\mathbf{s}_{1},...,\mathbf{s}_{n_s}\}$, it is a common practise to use the attention mechanism to generate $n_{s}$ weights to selectively attend the most relevant sentence. Then, the sentence features are aggregated to a fixed-length vector representation for a bag. 

In our work, the attention parameters will be generated by the dynamic parameter generator, and the attention weights are calculated as follows:
\begin{align}\label{eq6}
    &\overline{\mathbf{r}}_k = \phi_a(\mathbf{T}_{e}, \mathbf{r}_k), \nonumber \\
    &\alpha_{i} = \mathbf{s}_{i}\overline{\mathbf{r}}_k, \nonumber \\
    &\mathbf{z}_k = \sum_{i=1}^{n_{s}}\frac{\exp(\alpha_{i})}{\sum_{j=1}^{n_s} \exp(\alpha_{j}) }\mathbf{s}_{i},
\end{align}
where $\overline{\mathbf{r}}_k$ is dynamic attention parameters for the $k$-th class. $\mathbf{s}_{i}$ is the sentence feature. Note that we run the dynamic attention $n$ times to obtain $n$ aggregation results, i.e., $[\mathbf{z}_1,...,\mathbf{z}_n]$.

\subsubsection{Dynamic Classifier}

Each result $\mathbf{z}_k$ is classified by its corresponding classifier. In other words, the decision value for the $k$-th class is
\begin{align}
    &\overline{\mathbf{w}}_k = \phi_c(\mathbf{T}_{e}, \mathbf{w}_k), \nonumber \\
    &v = \overline{\mathbf{w}}_k\mathbf{z}_k + b_k,
\end{align}where $\overline{\mathbf{w}}_k$ is dynamic classifier parameters for the $k$-th class.
and $b_k$ is a bias term. Note that at the test stage, we do not know the ground-truth relation category $k$, thus we run the dynamic attention and the dynamic classifier $n$ times with a hypothesis $k$ each time. Each run will produce a posterior probability for the $k$-th class, and this result will be used for prediction and evaluation. The same operation has also been used in \cite{lin2016neural}.

\section{Experimental Results}

In this section, we first describe the dataset and evaluation criteria. Then, we show our implementation details. Finally, we report our results compared with other existing methods.

\subsection{Dataset}

We evaluate our method on a widely used dataset\footnote{http://iesl.cs.umass.edu/riedel/ecml/}, NYT, which is developed by \cite{riedel2010modeling}. The NYT dataset is generated by aligning Freebase \cite{bollacker2008freebase} relation facts with the New York Times corpus. The entities in sentences are recognized by the Stanford named entity tagger \cite{manning-etal-2014-stanford} and further matched the corresponding Freebase entities. The NYT dataset has been widely used as a benchmark in the existing literature \cite{hoffmann2011knowledge,surdeanu2012multi,zeng2015distant} etc. The sentences from the year 2005-2006 in New York Times corpus are used to generate the training set and the sentences from the year 2007 are used as the testing set. There are 53 relation classes in the dataset, including NA which means no relation between two entities.

\subsection{Evaluation Criteria}

Following the existing works \cite{mintz2009distant,lin2016neural}, we use a held-out evaluation method to evaluate the models. The held-out evaluation method compares the predicted relation classes with the ground truth to evaluate the corresponding method. The Precision-Recall (PR) curves and the top-N precision (P@N) will be reported for analysis. Moreover, to further evaluate our method on long-tail relations, we follow \cite{han2018hierarchical,katt_naacl19} and apply Hits@K metrics. In Addition, in the ablation study, we use AUC for quantitative analysis.

\begin{table}[htb]
\caption{\label{parameter_set} Hyper-parameter settings used in our model.}
\begin{center}
\begin{tabular}{lc}
\hline
\bf Parameter Name & \bf Value \\ 
\hline
Word dim. $d_{w}$ & 50  \\
Position dim. $d_{p}$ & 5 \\
Type dim. $d_{t}$ & 50 \\
Sentence emb. dim. $d_{s}$ & 690 \\
Relation emb. dim. $d_{r}$ & 690 \\
Window size $w$ & 3 \\
Batch size $B$ & 160 \\
Learning rate ${\lambda}_{lr}$ & 1.0 \\
Dropout probability $p_d$ & 0.5 \\
L2-regularization ${\lambda}_{L_2}$ & 1e-5 \\
\hline
\end{tabular}
\end{center}
\end{table}

\subsection{Implementation Details}

We use the same hyper-parameter settings in PCNN \cite{zeng2015distant}. The dimension of entity type and relation tuple element embedding are both set to 50. GCN layers are set to 2. The cross-entropy loss function is applied to train our model. The Adadelta optimizer \cite{zeiler2012adadelta} with its default parameters is used as the optimizer. Moreover, the dropout strategy \cite{srivastava2014dropout} is used at the classification layer, and L2-regularization is also used to prevent the model training from over-fitting. We implement the network based on PyTorch. Table \ref{parameter_set} reports all the hyper-parameter settings.

\begin{figure}[tb]
\begin{center}
\includegraphics[width=.8\textwidth]{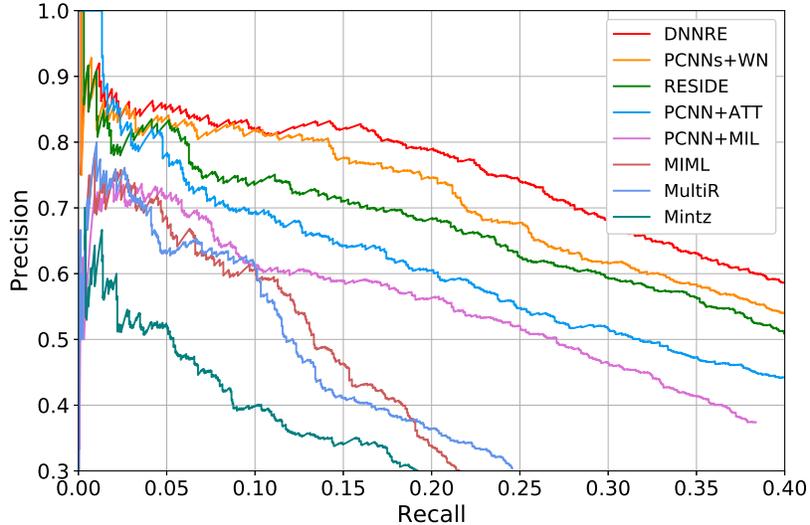}
\caption{\label{allpr}Performance comparison of proposed models with other methods.}
\end{center}
\end{figure}

\subsection{Overall Evaluation Results}

To evaluate the performance of \textbf{DNNRE}, we compare it against several existing hand-crafted feature based and deep-feature based methods, which are as follows:

%The hand-crafted feature based methods:
\begin{itemize}
\item \textbf{Mintz}: \cite{mintz2009distant} represents a traditional multi-class logistic regression DSRE model.
\item \textbf{MultiR}: \cite{hoffmann2011knowledge} proposes a probabilistic graphical model for multi-instance learning.
\item \textbf{MIML}: \cite{surdeanu2012multi} applies a graphical model which jointly models multiple instances and multiple labels.
\item \textbf{PCNN+MIL}: \cite{zeng2015distant} is a convolutional neural network (CNN) adopting a piecewise max-pooling method for sentence representation. 
\item \textbf{PCNN+ATT}: \cite{lin2016neural} uses selective attention over instances to aggregate sentence embeddings to a bag-level embedding, which can utilize the information in other sentences.
\item \textbf{RESIDE}: \cite{vashishth2018reside} utilizes side information (i.e., relation alias and entity type information) and GCN is used to encode syntactic information of instances for boosting the DSRE performance. 
\item \textbf{PCNNs+WN} \cite{yuan2019distant} proposes to use a linear attenuation simulation to weight the word embeddings and a non-independent and identically distributed relevance hypothesis is used to capture the relevance of sentences in the bag.
\end{itemize}

From the PR curves in Figure \ref{allpr}, it can be observed that DNNRE achieves superior performance compared with the state-of-the-arts. The precision value of DNNRE outperforms others under almost all recall values. Especially, when recall ranges from 0.10 to 0.40, there is a consistent margin between DNNRE and other methods. By cross-referencing the P@N results in Table \ref{Precision@N}, it is clear that our method achieves a significant improvement over the comparing methods. 

To highlight, compared with a recent method RESIDE, which utilizes side information (e.g., entity type and relation alias).
DNNRE achieves higher precision by a large margin over all recall values in PR curves and attains an improvement pf 4.2\% in P@N on average. That demonstrates that our method of using the entity type information is more effective.

The performance of DNNRE indicates that the design of the dynamic network can take advantage of the class relation with corresponding entity types. It can dynamically adapt its parameters to represent the relations more accurately. A case study is reported for evaluating the effectiveness of DNNRE for the style shift problem caused by keyword variation in subsection \ref{case_study}.

\begin{table}[tb]
\caption{\label{Precision@N} P@N comparison of proposed models with other methods. The best results are in bold font.}
\begin{center}
\begin{tabular}{cccccc}
\hline 
P@N & PCNN+MIL & PCNN+ATT & RESIDE & PCNNs+WN & DNNRE \\
\hline 
P@100 & 72.3 & 76.2 & 84.0 & 83.0 & \bf85.0 \\

P@200 & 69.7 & 73.1 & 78.5 & 82.0 & \bf 83.0 \\

P@300 & 64.1 & 67.4 & 75.6 & 80.3 & \bf 82.7 \\
 
Mean & 68.7 & 72.2 & 79.4 & 81.8 & \bf 83.6 \\
\hline
\end{tabular}
\end{center}
\end{table}

\begin{table}[tb]
\caption{\label{HitsK}Accuracy (\%) of Hits@K on relations with training instances fewer than 100/200.}
\begin{center}
%\resizebox{0.46\textwidth}{11.5mm}{
\begin{tabular}{ccccccc}
\hline 
\# Instances & \multicolumn{3}{c}{$<100$} & \multicolumn{3}{c}{$<200$}\tabularnewline
Hits@K  & \multicolumn{1}{c}{10} & \multicolumn{1}{c}{15} & 20 & \multicolumn{1}{c}{10} & \multicolumn{1}{c}{15} & 20\tabularnewline
\hline 
PCNN+ATT & $<5.0$ & 7.4 & 40.7 & 17.2 & 24.2 & 51.5\tabularnewline
PCNN+HATT & 29.6 & 51.9 & 61.1 & 41.4 & 60.6 & 68.2\tabularnewline
PCNN+KATT & 35.3 & \textbf{62.4} & 65.1 & 43.2 & 61.3 & 69.2\tabularnewline
DNNRE & \textbf{57.6} & 62.1 & \textbf{66.7} & \textbf{64.1} & \textbf{68.0} & \textbf{71.8}\tabularnewline
\hline 
\end{tabular}{\scriptsize\par} %}
\end{center}
\end{table}

\subsection{Evaluation for Long-tail Relations}

We also evaluate the performance of DNNRE on \textbf{Long-tail Relations} by following the protocol of \cite{han2018hierarchical,katt_naacl19}: (1) A subset of the test dataset in which all the relations have fewer than 100/200 training instances is selected. (2) Hits@K with $K=\left\{10,15,20\right\}$ metrics is used as an evaluation metric, which measures the likelihood of true relation falls into the first K candidate relations recommended by the model. 

In Table \ref{HitsK}, it is observed that our method outperforms \textbf{PCNN+ATT} \cite{lin2016neural}, \textbf{PCNN+HATT} \cite{han2018hierarchical} and \textbf{PCNN+KATT} \cite{katt_naacl19} in most of the Hits@K. In particular, we observe that DNNRE achieve 57.6\% and 64.1\% accuracies for Hits@10. That is, more than half of true classes of the long-tail samples can be predicted into the first $10$ candidate relation recommended by DNNRE, which outperforms other works significantly at least 20\% absolute improvement.
This demonstrates that DNNRE can substantially boost the performance of long-tail relation classes.

\section{Analysis and Discussion}
In this section, we first conduct the ablation study to analyze the effects of each dynamic component of DNNRE (Subsection {\ref{ablation_study_over}). 
Secondly, we compare our relation-aware attention over entity types with other aggregation strategies, i.e., average pooling and max pooling (Subsection {\ref{aggregation_strategy}). 
Then, we investigate the effects of the different granularities of the entity types on the capability of the generated dynamic parameters (Subsection \ref{diff_grained}).
Finally, a case study is given to demonstrate the effectiveness of DNNRE (Subsection \ref{case_study}).

\begin{figure}[htb]
\centering
\includegraphics[width=.8\linewidth]{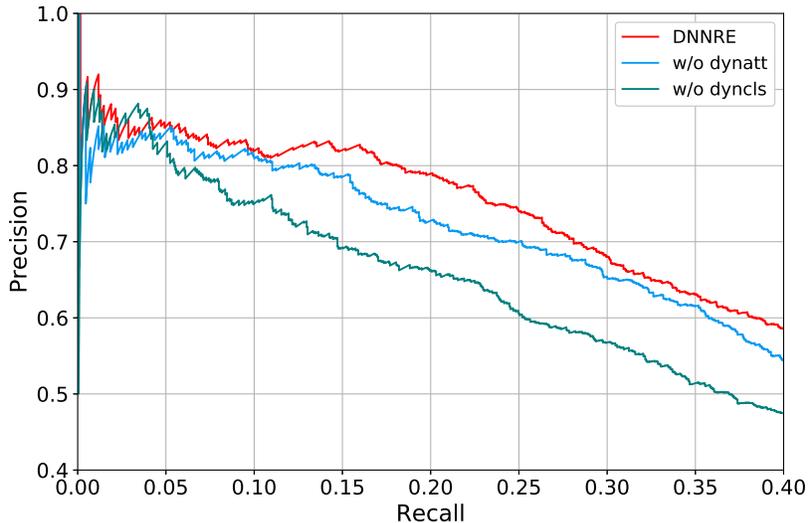}
\caption{Ablation study of different dynamic components of DNNRE.}
\label{ablation}
\end{figure}

\subsection{Ablation Study} \label{ablation_study_over}

In this subsection, we conduct ablation studies to validate the effect of each component of DNNRE (i.e., the dynamic attention and dynamic classifier). 
The applied evaluation metrics we use are the PR curve and AUC. Note that since some bags in the testing set are noisy, we use AUC (recall $<0.4$) to focus on the high confidence bags in the low-recall region. 

In Figure \ref{ablation}, we report different ablated version of DNNRE, which is described as follows:
\begin{itemize}
\item \textbf{w/o dynatt} denotes a variant by removing the dynamic attention and only use the vanilla static attention parameters.
\item \textbf{w/o dyncls} denotes a variant by removing the dynamic classifier and only use the vanilla static classifier parameters.
\end{itemize}
Merely using dynamic attention (\textbf{w/o dyncls}) or dynamic classifier (\textbf{w/o dynatt}) can boost the AUC to around 0.264 and 0.290, respectively. Particularly, when the dynamic classifier is removed, the performance drops to 0.264. As a result, it indicates that the dynamic classifier is more important for the DNNRE. 
Besides, when these two dynamic components are both utilized in our model, the DNNRE achieves an AUC of 0.303, which means that the dynamic attention and classifier are complementary to each other. The results demonstrate that each dynamic module contributes to the superior performance of DNNRE, that is, both the dynamic design of the attention and classifier are beneficial for relation recognition.

\begin{figure}[htb]
\begin{center}
\includegraphics[width=.8\textwidth]{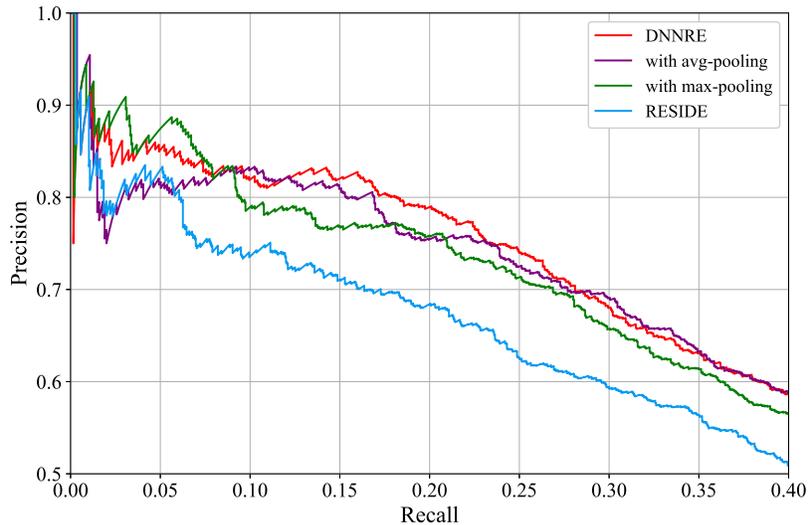}
\caption{\label{aggregation_pr}Performance comparison of different aggregation strategies.}
\end{center}
\end{figure}

\subsection{Effects of Different Aggregation Strategies} \label{aggregation_strategy}

In this subsection, we investigate the effects of different aggregation strategies. 
Besides the relation-aware attention over entity types, we also apply another two aggregation strategies, i.e., average pooling and max pooling.

As shown in Figure \ref{aggregation_pr}, \textbf{with avg-pooling} and \textbf{with max-pooling} denote the relation-aware attention is replaced with the average pooling operation and the max pooling operation, respectively. It is observed that the proposed relation-aware attention over the entity types is more robust in the PR curves which indicates that this attention mechanism is capable of selecting the important information from entity types for different relation classes.

Note that we also report the PR curve of RESIDE \cite{vashishth2018reside} in Figure \ref{aggregation_pr}. The performance of DNNRE all outperforms RESIDE by a large margin with any chosen aggregation strategies, which demonstrates that the dynamic design of our model using the entity type information is more effective.

\begin{figure}[htb]
\begin{center}
\includegraphics[width=.8\textwidth]{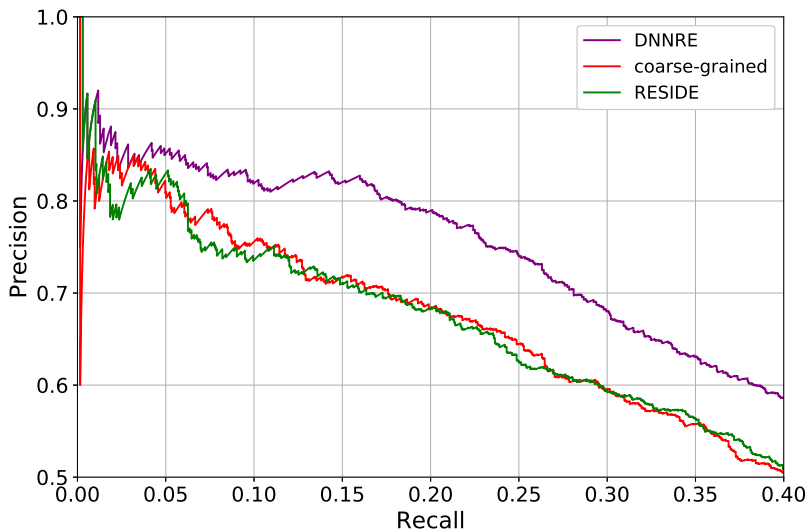}
\caption{\label{coarse_pr} Performance comparison of the  DNNRE with different granularities of entity types.
}
\end{center}
\end{figure}

\subsection{Effects of Different Granularities of Entity Type} \label{diff_grained}

In this subsection, we investigate how the different granularities of the entity types affect the discriminative capability of the generated dynamic parameters. The results are reported in Figure \ref{coarse_pr}, in which the \textbf{coarse-grained} denotes that the dynamic parameters of the attention and classifier are generated by the coarse-grained entity types, i.e., the 38 coarse types which form the first hierarchy of FIGER types.

On the one hand, DNNRE with fine-grained entity types outperforms the DNNRE with \textbf{coarse-grained} entity types by a large margin. That means the fine-grained entity type is helpful for the performance improvement of DSRE, which is also validated by \cite{liu2014exploring}. 

On the other hand, considering RESIDE \cite{vashishth2018reside} also utilizes the 38 coarse types of FIGER \cite{ling2012fine}, we compare the DNNRE (\textbf{coarse-grained}) with RESIDE. Note that RESIDE utilizes lots of external information, including the entity type, relation alias and syntactic information of sentence, which is encoded by a Graph Convolutional Network \cite{kipf2017semi}. However, only using the 38 coarse types, the performance of our DNNRE is still comparable to RESIDE, which indicates that the way of generating dynamic parameters by the entity types is more effective than utilizing them as input constraints \cite{liu2014exploring,vashishth2018reside}.

\begin{figure}[htb]
\begin{center}
\includegraphics[width=1.\textwidth]{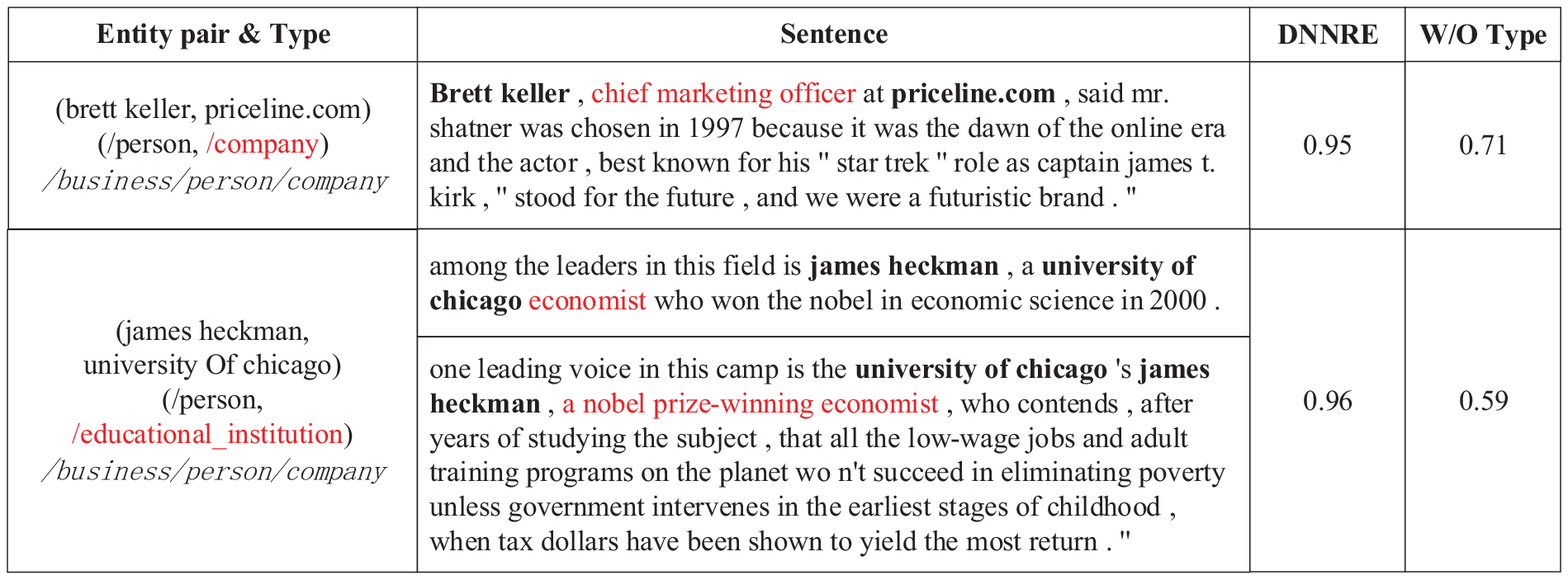}
\caption{\label{case} Examples to evaluate DNNRE for the style shift problem. On the left side, the entity pairs, entity types, and relation classes of two bags are shown. On the right side, the estimated probabilities (confidence scores) for detecting the ground-truth relation are shown.
}
\end{center}
\end{figure}

\subsection{Case Study} \label{case_study}

Figure \ref{case} uses two examples to show how DNNRE addresses the style shift problem. Two models are used to conduct the case study, i.e., DNNRE and its variant (\textbf{W/O Type}), whose parameters are not incorporating the entity type information.

The first example expresses the relation by the keyword ``\textit{chief marketing officer}''. However, in the second example, the tail entity type changes to ``\textit{educational institution}'', and the sentence expresses the same relation by a different keyword, i.e., ``\textit{economist}''. 
In both examples, the proposed DNNRE is able to adjust the model parameters according to entity type information and produce higher prediction scores. The confidence scores from DNNRE are 0.95 and 0.96 for these two examples, respectively. On the contrary, the confidence scores from DNNRE \textbf{W/O type} only obtain 0.71 and 0.59, respectively. The results demonstrate that DNNRE can utilize the entity type information to compensate for the relation representations with more varieties for handling the style shift problem.

\section{Conclusion}

In this work, we propose a novel Dynamic Neural Network for Relation Extraction (DNNRE), whose parameters are determined by the query entity types and relation classes. The dynamic design of our model benefits for the potential style shift caused by keyword variation under different entity types. Besides, the entity type information can also be trained cross different relation classes and further benefits for the long-tail relation classes.
Through extensive experiments, we demonstrate that the proposed method is effective for improving the DSRE accuracy.

In future work, we will focus on the following directions: (1) We will explore how to better model the style shift problem with different information inputs, e.g., the entity description information. (2) The connections between different relation classes can be further utilized to boost the performance of DSRE. (3) Entity typing can be incorporated into our DNNRE via a multi-task training manner and further benefits for the improvement of the DSRE performance.

\section*{Acknowledgment}
This work was supported by the Key Research and Development Program of Sichuan Province (2019YFG0409).

\bibliographystyle{elsarticle-num} 
\bibliography{mybibfile}

\end{document}